\documentclass{article}

\usepackage{algpseudocode}
\usepackage{algorithm}
\usepackage{graphicx}
\usepackage[english]{babel}
\usepackage{amsmath, amsfonts}
\usepackage{authblk}
\usepackage{subcaption}
\usepackage{adjustbox}
\usepackage[colorlinks,allcolors=blue]{hyperref}

\usepackage{lineno,hyperref}

\title{Optimizing Convolutional Neural Network Architecture}
\author[1]{Luis Balderas
	\thanks{Corresponding author. luisbalru@decsai.ugr.es}}

\author[2]{Miguel Lastra}

\author[1]{Jose M. Benítez}

\affil[1]{Department of Computer Science and Artificial Intelligence, DiCITS, iMUDS, DaSCI, E.T.S.I.I.T. University of Granada, Spain}
\affil[2]{Department of Software Engineering, DiCITS, iMUDS, DaSCI, E.T.S.I.I.T. University of Granada, Spain}

\begin{document}
	
	\maketitle
	
	\begin{abstract}
		Convolutional Neural Networks (CNN) are widely used to face challenging tasks like speech recognition, natural language processing or computer vision. As CNN architectures get larger and more complex, their computational requirements increase, incurring significant energetic costs and challenging their deployment on resource-restricted devices. In this paper, we propose Optimizing Convolutional Neural Network Architecture (OCNNA), a novel CNN optimization and construction method based on pruning and knowledge distillation designed to establish the importance of convolutional layers. The proposal has been evaluated though a thorough empirical study including the best known datasets (CIFAR-10, CIFAR-100 and Imagenet) and CNN architectures (VGG-16, ResNet-50, DenseNet-40 and MobileNet), setting Accuracy Drop and Remaining Parameters Ratio as objective metrics to compare the performance of OCNNA against the other state-of-art approaches. Our method has been compared with more than 20 convolutional neural network simplification algorithms obtaining outstanding results.  As a result,  OCNNA is a competitive CNN constructing method which could ease the deployment of neural networks into IoT or resource-limited devices.
	\end{abstract}
	
	\section{Introduction}
	Over the last years, deep neural networks (DNN) have become the state-of-art technique on several challenging tasks such as speech recognition \cite{sp-rec}, natural language processing \cite{nlp} or computer vision \cite{10.1145/3065386}. In particular, Convolutional Neural Networks (CNNs) have achieved an extraordinary success in a variety range of computer vision problems, such as image classification \cite{10.1145/3065386}, object detection in images \cite{ZHAO2021107796}, object detection in video \cite{OLMOS201866}, semantic segmentation \cite{YU2021107756}, video restoration \cite{10.5555/3298023.3298157} or medical diagnosis \cite{cai2020}. The astonishing results of CNNs are associated with the huge amount of annotated data and important advances in hardware. Unfortunately, CNNs usually have an immense number of parameters, incurring in high storage requirements, significant computational and energetic costs \cite{han2015} and, as a result, an evident environmental impact. In 2019, researchers at the University of Massachusetts found that training several deep learning models (including neural architecture search) can emit more than 284019 kilograms of carbon dioxide, in other words, nearly five times the lifetime emissions of the average American car (including the manufacture of the car itself) \cite{str2019}. Concretely, to classify a single image, the VGG-16 model \cite{https://doi.org/10.48550/arxiv.1409.1556} requires more than 30 billion floating-point operations (FLOPs) and contains 138 million parameters requiring more than 500 MB of space \cite{ALQAHTANI2021103220}. In consequence, reducing the model's storage requirements and computational cost becomes critical for resource-limited devices, specially in IoT applications, embedded systems, autonomous agents, mobile devices or edge devices \cite{8703045}.
	
	Nonetheless, \cite{10.5555/2968826.2968968} remarks that a main DNN's property is their considerable redundancy in parameterization, which leads to the idea of reducing this redundancy by compressing the networks. However, a severe problem along with compressing the models is the loss of accuracy. In order to avoid it, there are some ways of designing efficient DNNs and generating effective solutions. For example, using memetic algorithms to find a good architecture to fit the task or, provided an architecture, using alternative optimization algorithms to fine-tune the connection weights. Kernel Quantization is also used for efficient network compression \cite{9672186}. Alternatively, pruning is one of the most used methods to reduce neural networks' complexity. In fact, pruning techniques have been extensively studied for model compression since 1990, when Optimal Brain Damage (OBD) \cite{OBD} and Optimal Brain Surgery (OBS) \cite{OBS} where designed. Along the past years, many other approaches have been presented in order to generate more efficient and effective neural networks in all their representations (e.g. dense, convolutional or recurrent). Obtaining a subnetwork with far fewer parameters without compromising accuracy is the main goal of pruning algorithms.
	
	In this paper, we propose a novel CNN optimization and construction technique called Optimizing Convolutional Neural Network Architecture (OCNNA) based on pruning and knowledge distillation \cite{kd} which requires minimal tuning. OCNNA has been designed to assess the importance of convolutional layers. Since this measure is computed for every convolutional layer and unit from the model output to the input, it essentially reflects the importance of every single convolutional filter and its contribution to the information flow through the network. Moreover, our method is able to order convolutional filters within a layer by importance and, as a consequence, it can be seen as a feature importance computation tool which can generate more explainable neural networks. OCNNA is easy to apply, having only one parameter $k$, called percentile of significance, which represents the proportion of filters which will be transferred to the new model based on their importance. Only the $k$-th percentile of filters  with higher values after applying OCNNA process will remain. The proposed OCNNA can directly be applied to trained CNNs, avoiding the training process from scratch.
	
	We have thoroughly evaluated the optimization efficacy of the OCNNA method compared to the state-of-art pruning techniques. Our experiments on CIFAR-10, CIFAR-100 \cite{cifar} and Imagenet \cite{imagenet} datasets and for popular CNN architectures such as ResNet-50, VGG-16, DenseNet40 and MobileNet show that our algorithm leads to better performance in terms of the accuracy in prediction and the reduction of the number of parameters.
	
	The rest of this paper is structured as follows: In Section 2, we introduce the state-of-art of different approaches for designing efficient deep neural networks. In Section 3, we describe our proposal. In Section 4 our methodology is experimentally analyzed. In Section 5 we discuss the results and Section 6 highlights the conclusions. 	
	\section{Previous work}
	
	The purpose of this section is to make a brief overview of the main approaches of model compression: Neuroevolution, Neural Architecture Search, Quantization and Pruning. Our research is mainly focusing on convolutional neural network pruning.
	
	\subsection{Neuroevolution}
	
	Neuroevolution can be applied in several tasks related with efficient neural network design, such as learning Neural Networks (NN) building blocks, hyperparameters or architectures. In 2002, NeuroEvolution of Augmenting Topologies (NEAT) was presented in \cite{Stanley2002EvolvingNN}, showing the effectiveness of a Genetic Algorithm (GA) to evolve NN topologies and strengthening the analogy with biological evolution. More recently, \cite{10.5555/3305890.3305981} took inspiration from NEAT and evolved deep neural networks by starting with a small NN and adding complexity through mutations. In \cite{Real_Aggarwal_Huang_Le_2019} a more accurate approach can be found, which consists in stacking the same layer module to make a deep neural network, like Inception, DenseNet and ResNet \cite{Stanley2019}.

	\subsection{Neural Architecture Search}
	Recently, Neural Architecture Search (NAS), whose main goal is to automatically design the best DNN architecture, has achieved a great importance in model design. On one hand, NAS algorithms can be divided into two categories: (1) Microstructure search: searching the optimal operation for each layer; (2) Macrostructure search: searching the optimal number of channels/filters for each layer or the optimal depth of the model \cite{WANG2021533}. On the other hand, based on the optimizer used, the existing NAS algorithms can be classified into three categories: reinforcement learning based NAS algorithms, gradient-based NAS algorithms and evolutionary NAS algorithms (ENAS). In this sense, NSGA-II has been recently used for NAS creating NSGA-Net \cite{10.1145/3321707.3321729}. In \cite{9336247}, NATS-BEnch is proposed, consisting in a unified benchmark for topology and size aggregating more than 10 state-of-the-art NAS algorithms.

	\subsection{Quantization}
	
	Quantization is known as the process of approximating a continuous signal by a set of discrete symbols or integer values \cite{LIANG2021370}. In other words, it reduces computations by reducing the precision of the datatype. Advanced quantization techniques, such as asymmetric quantization \cite{8578384} or calibration based quantization, have been presented to improve accuracy. In \cite{LIANG2021370} we find a complete Quantization guide and recommendations. 
	
	\subsection{Knowledge Distillation}
	
	Knowledge Distillation, which was first defined by \cite{10.1145/1150402.1150464} and generalized in \cite{https://doi.org/10.48550/arxiv.1503.02531} is the process of transferring knowledge from one neural network to a different one. In \cite{9340578} a Student-Teacher framework is presented, introducing different scenarios such as distillation based on the number of teachers (from one teacher vs. multiple teachers), distillation based on data format (data-free, with a few samples or cross-modal distillation) or online and teacher-free distillation. In this work, an alternative type of knowledge distillation is proposed through the application of OCNNA, which consists of transferring knowledge from the original model to the optimized model by sharing the weights of useful convolutional units. Thus, all the knowledge generated thanks to the original CNN training phase is transferred to the simplified model, disregarding those units that contribute less useful information in prediction.
	 
	\subsection{Pruning}
	Network pruning is one of the most effective and prevalent approaches for model compression. Pruning techniques can be classified by various aspects: structured and unstructured pruning, depending on whether the pruned network is symmetric or not \cite{https://doi.org/10.48550/arxiv.1810.05270}; neuron, weight, or filter pruning depending on the network's element which is pruned; or static and dynamic pruning. While static pruning removes elements offline from the network after training and before inference, dynamic pruning determines at runtime which elements will not participate in further activity \cite{LIANG2021370}. Most researchers focus on how to find unimportant filters. Magnitude-based methods \cite{8237560} take the magnitude of weights of feature maps from some layers as the importance criterion and prune those with small magnitude. Others measure the importance of a filter through their reconstruction loss (Thinet) \cite{8416559}  or Taylor expansion \cite{https://doi.org/10.48550/arxiv.1611.06440}, \cite{qi2021}. In \cite{https://doi.org/10.48550/arxiv.1607.03250}, Average Percentage of Zeros (APoZ) is presented to measure the percentage of zero activations of a neuron after the ReLU mapping, pruning the redundant ones. HRank \cite{https://doi.org/10.48550/arxiv.2002.10179} understands filter pruning as an optimization problem, using the feature maps as the function which measures the importance of a filter part of the CNN. In \cite{WANG202141}, a new CNN compression technique is presented based on pruning filter-level redundant weights according to entropy importance criteria (FPEI) with different versions depending on the learning task and the NN. SFP \cite{10.5555/3304889.3304970} and FPGM, based on filter pruning via geometric median \cite{8953212}, use soft filter pruning; PScratch \cite{https://doi.org/10.48550/arxiv.1909.12579} proposes to prune from scratch, before training the model. In \cite{YEOM2021107899} a criterion for CNN pruning inspired by NN interpretability is proposed: the most relevant units are found using their relevance scores obtained from concepts of explainable AI (XAI). \cite{ARADHYA2022638} introduces a data-driven CNN architecture determination algorithm called AutoCNN which consists of three training stages (spatial filter, classification layers and hyperparameters). AutoCNN uses statistical parameters to decide whether to add new layers, prune redundant filters, or add new fully connected layers pruning low information units. An iterative pruning method based on deep reinforcement learning (DRL) called Neon, formed by a DRL agent which controls the trade-off between performance and the efficiency of the pruned network, is introduced in \cite{HIRSCH2022381}. For each hidden layer, Neon extracts the architecture-based and the layer-based feature maps which represent an observation. Then, the afore-mentioned hidden layer is compressed and fine-tuned. After that, a reward is calculated and used to update the  deep reinforcement learning agent's weights. This process is repeated several times for the whole neural network. In \cite{FERNANDESJR202129}, a multi-objective evolution strategy algorithm, called DeepPruningES is proposed. Its final output will be three neural networks with different trade-offs called knee (with the best trade-off between training error and the number of FLOPs), boundary heavy (with the smallest training error) and boundary light solutions (with the smallest number of FLOPs). In \cite{WANG2021533}, a customized correlation-based filter level pruning method for deep CNN compression, called COP, is presented, removing redundant filters through their correlations. Finally, in \cite{LI202282} SCWC is introduced, a shared channel weight convolution approach to reduce the number of multiplications in CNNs by the distributive property due to the structured channel parameter sharing.	
\section{Proposal}

In this paper, we address the challenge of establishing the topology of a convolutional neural network. Thus, we introduce the Optimizing Convolutional Neural Network Architecture (\textbf{OCNNA}), a new convolutional neural network construction method based on pruning and knowledge distillation. In this section we pay special attention to the process that performs the identification and extraction of significant filters.

\subsection{Notation}

First of all, we introduce some symbol notations used throughout the article. Suppose we have a deep neural network with $L$ convolutional layers. Let $w_m^{l}$ and $o_m^{l}$ be the convolutional filter and the output of the $l$-th layer. The subscript $m \in 1,\dots, M^l$ represents the filter index, where $M^l$ indicates the total number of output filters in the corresponding layer. In consequence, pruning the $l$-th filter in layer $m$ implies removing the corresponding $w_m^{l}$.

Principal Component Analysis (PCA) \cite{kurita2019principal} is a data analysis tool applied to identify the most meaningful basis to reexpress, revealing a hidden structure, a given dataset. We will define $P_m^{l} = PCA(o_m^{l})$ as the matrix result of computing PCA on the $m$-th filter's output of the $l$-th layer.

The Frobenius norm \cite{fn} is a norm of an $m\times n$ matrix $A$ defined as
\begin{equation}
	||A||_{F} = \sqrt{\sum_{i=1}^{m} \sum_{j=1}^{N} |a_{ij}|^2} \label{fn1}
\end{equation}

It is also equal to the square root of the matrix trace of $A, A^H$

\begin{equation}
	||A||_{F} = \sqrt{Tr(A A^H)}  \label{fn2}
\end{equation}

\noindent where $A^H$ is the conjugate transpose \cite{mathworld}. Let's define $F_m^{l} = ||P_m^{l}||_{F}$ as the Frobenius norm of the above PCA calculation (on the $m$-th filter's output of the $l$-th layer).

The Coefficient of Variation (CV) is the relationship between mean and standard deviation \cite{everitt1998}. If $D$ is a data distribution, $\sigma_D$ its standard deviation and $\mu_D$ its mean, CV is calculated as

\begin{equation}
	CV_D = \frac{\sigma_D}{\mu_D} \label{fn3}
\end{equation}
It is attractive as a statistical tool because it permits the comparison of variables free from scale effects (dimensionless). We will call $C = CV(x)$ if $x \in \mathbb{R}^n$, for a given $n \in \mathbb{N}$.

Finally, we define the percentile of significance $k$. Only the $k$-th percentile of filters with higher values in terms of importance will remain. All notations can be found in Table \ref{tab:not}.

\begin{table*}[!t]
	\caption{Notations and definitions\label{tab:not}}
	\centering
	\begin{tabular}{|c|c|}
		Notation & Definition \\
		$L$ & Number of convolutional layers\\
		$w_m^{l}$ & $m$-th filter from the $l$-th convolutional layer \\
		$o_m^{l}$ & output of the $m$-th filter from the $l$-th  layer \\
		$M^l$ & number of output filters in the $l$-th  layer \\
		$P_m^l$ & PCA applied to the $m$-th filter from the $l$-th  layer \\
		
		$F_m^l$ & Frobenius norm of $P_m^l$ \\
		$C$ & Coefficient of Variation of a vector \\ 
		$k$-th percentile & Percentile of significance  \\
	\end{tabular}
\end{table*}

\subsection{OCNNA: the algorithm}

\begin{figure*}[!t]
	\centering
	\includegraphics[width=\textwidth]{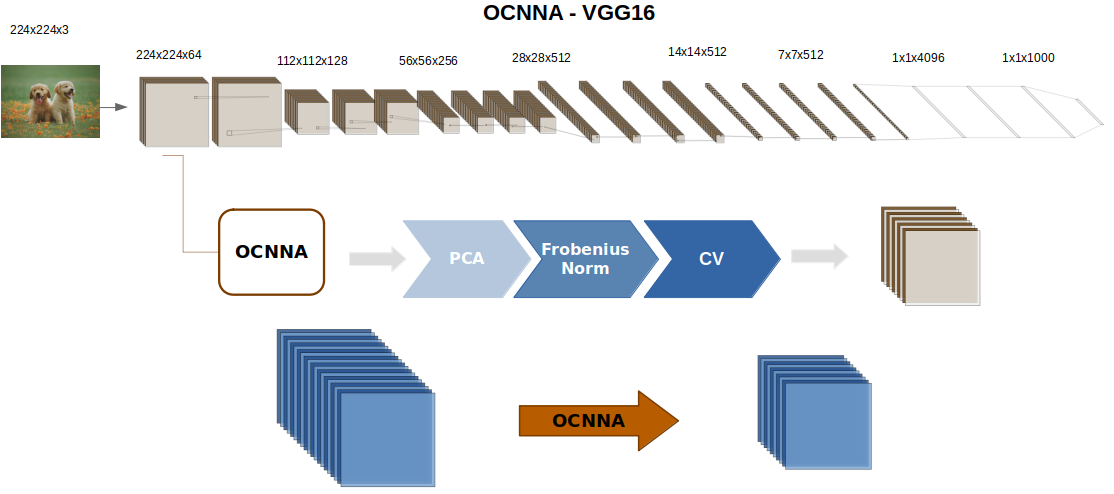}
	\caption{OCNNA applied to VGG-16. Given the output from  the $i$-th convolutional layer evaluated in a validation set, PCA, Frobenius norm and Coefficient of Variation are applied in order to measure the most significant filters. The $k$-th percentile of filters, in terms of importance, are selected, generating a new model which $i$-th convolutional layer is a optimized version of the original one. This approach is applied to every convolutional filter.}
	\label{ocnna_vgg16}
\end{figure*}

OCNNA is designed to identify the most important convolutional filters in a CNN, creating a new model in which the less significant convolutional units are not included. In consequence, OCNNA generates a simpler model in terms of the number of parameters with minimum precision loss, as we will see in the next section. Besides, OCNNA allows to order the convolutional filters by importance, providing a feature importance method and, as a result, helping to create more explainable IA models.  Our method employs three important techniques to identify the significant filters: Principal Components Analysis (PCA), for selecting the most important features based on their hidden structure, the Frobenius norm, summarizing the PCA output information, and the Coefficient of Variation (CV), measuring the variability of Frobenius norm outputs. Figure \ref{ocnna_vgg16} shows the simplification process of a VGG-16 convolutional filter.  We have thoroughly evaluated our CNN building scheme for the ResNet-50 architecture too. Algorithm 1 presents OCNNA. The essential part of our method is the criterion of filter importance.

Given the  convolutional layer $w^l$, a $D_{var}$ dataset, used exclusively for measuring the importance of filters, is evaluated generating the output $o^l$. $o^l$ is formed by $M^l$ filters. In consequence, $o_m^l$ is the $m$-th filter's output from the $l$-th layer. OCNNA is applied to $o_l$ in order to measure the importance the $M^l$ filters importance in the $m$-th layer. Concretely,  we adopt a three-stage process. First, for each filter and image contained in $D_{var}$, we apply PCA retaining the $95\%$ of variance.

\begin{equation}
	P_m^l = PCA(o_m^l) \label{pca}
\end{equation}

\noindent with $P_m^l$ a matrix which contains the most meaningful features generated by the $m$-th filter of the $l$-th convolutional layer. Nonetheless, the information embedded in $P_m^l$ is too large. In consequence, we apply the Frobenius Norm to summarize this information:

\begin{equation}
	F_m^l = ||P_m^l||_F \label{fn}
\end{equation}

\noindent obtaining a vector $F_m^l$, in which each component is the result of the process described above applied to each image from $D_{var}$. Finally, we calculate the CV:

\begin{equation}
	C_m = CV(F_m^l) \label{cv}
\end{equation}

\noindent $C_m$ is a number which summarizes the $m$-th filter significance within the $l$-th convolutional layer by measuring the variability of the process PCA and Frobenius norm for each image in $D_{var}$. In other words, OCNNA gives a low score of importance to a filter if, for a bunch of images, generates an output whose hidden structures (PCA, 95\% variance), after being summarized (Frobenius norm), have little variability (CV).

\begin{figure*}[!t]
	\centering
	\includegraphics[width=\textwidth]{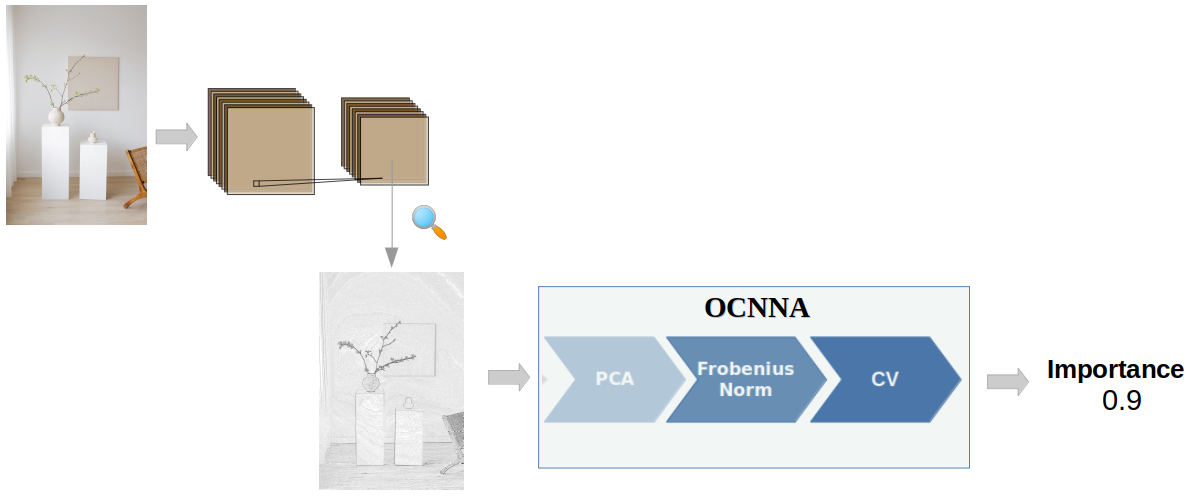}
	\caption{Application of OCNNA to one convolutional filter. As we can see, the filter generates partial information compared to the whole layer output. Applying OCNNA to the filter's output, our method provides a single number for this filter which reflects its importance. This process is iterated over all filters from a layer and the $k$-th percentile of them in terms of significance will form part of the new model.}
	\label{ocnna-filtro}
\end{figure*}

To sum up, OCNNA is able to extract insights and measure the importance of each filter of a convolutional layer, starting from hundreds of arrays which constitute the output from a dataset, called $D_{var}$,  and generating a holistic vision summarizing the filter significance into a single number (Figure \ref{ocnna-filtro}). As a result, OCNNA transforms the output of a convolutional layer into an array in which the $i$-th component represents the $i$-th filter's importance. Finally, using the parameter $k$, or percentile of significance, we extract the $k$-th percentile of filters in terms of significance, completing the simplification process. The larger is $k$, the more strict is the filter selection. In consequence, less filters will be selected and the new model will be simpler (less number of parameters compared to the original one). 

\begin{algorithm}
	\caption{OCNNA}
	\begin{algorithmic}[1]
		\Function{OCNNA}{model, $k$, $D_{var}$}
		\State opt\_model = Model()
		\For{layer in model.Layers do}
		\If{layer is Convolutional}
		\State variability = List()
		\For{filter in layers}
		\State $o = $ model.predict($D_{var}$)
		\State $p = $ PCA($o$, $95\%$ var)
		\State $pn = $ FrobeniusNorm($p$)
		\State $c = $ CV($pn$)
		\State variability.append($c$)
		\EndFor
		\State new\_layers\_index = get $k$-th percentile in variability
		\State Add new layer with new\_layers\_index filters from model
		\Else
		\State Add layer to opt\_model
		\EndIf
		\EndFor
		\State \Return opt\_model 
		\EndFunction
	\end{algorithmic} 
\end{algorithm}

\subsection{Implementation}

As we have said, OCNNA can measure the importance of a convolutional filter extracting hidden insights from a multidimensional array and express it through a number. This process, which must be completed for each image of a validation dataset, implies heavy computational costs. In this sense, OCNNA is designed to maximize its performance in terms of the time required to complete the simplification process by proposing a parallel computing paradigm. In other words, counting the number of CPUs available and distributing the tasks associated to each filter (prediction, PCA, Frobenius Norm and CV) of the convolutional layer among them, carrying out the calculations simultaneously and, as a result, speeding-up the results. This parallelism is absolutely transparent to the user. Once all the operations are finished, a synchronization process between the different subtasks is accomplished, mapping every result (the significance of the filter) in the correct component of the vector which represents the importance of each filter in the convolutional layer. It has been implemented in Python 3.9 and Tensorflow 2.9 has been used as machine learning framework.

\section{Empirical Evaluation}

To assess the performance of OCNNA, we have designed a thorough empirical procedure that includes different well known datasets (CIFAR-10, CIFAR-100 and Imagenet) and architectures (ResNet-50, VGG-16, DenseNet-40 and MobileNet) which represent a core benchmark extensively referenced in the literature. Moreover, OCNNA has been compared with 20 state-of-art CNN simplification techniques obtaining successful results. This section is structured as follows: The architectures and datasets, metrics, compared state-of-art approaches and training process settings are explained in order to assure the experiments' reproducibility. Finally, results and analysis for CIFAR and Imagenet datasets are presented comparing them with the other state-of-art techniques.

\subsection{Common architectures and datasets}

We have thoroughly evaluated our CNN building and optimizing scheme. In order to obtain comparable results with other state-of-art approaches, we have selected two popular CNN architectures: VGG-16 \cite{https://doi.org/10.48550/arxiv.1409.1556}, ResNet-50 \cite{https://doi.org/10.48550/arxiv.1512.03385}, DenseNet-40 \cite{https://doi.org/10.48550/arxiv.1608.06993} and MobileNet \cite{howard2017mobilenets}. VGG-16 is Convolutional Neural Network formed by $138.4$M of parameters and 16 layers. The input is a fixed-size $224 \times 224$ RGB image, which is passed through a stack of $3\times 3$ receptive field convolutional layers, with padding fixed to 1 pixel. Five max-pooling layers ($2\times2$ pixel window, stride 2), which follow some of the convolutional layers are included as spatial pooling. The last stack of convolutional layers is followed by three dense layers: 4096, 4096 and 1000 channels respectively. ResNet-50, inspired by the philosophy of VGG nets, introduces the concept of residual learning to ease the training process by reformulating the layers as learning residual functions with reference to the layer inputs, instead of learning unreferenced functions. In practice, ResNet contains shortcut connections. As a result, ResNet-50 has fewer filters and lower complexity than VGG-16, formed by $25.6$M of parameters. DenseNet ($1$M of parameters and 40 layers) connects each convolutional unit as it was a feed-forward neural network, reducing the number of  parameters and diminishing problems such as vanishing-gradient. Finally, MobileNet is a light weight neural network designed for mobile and embedded vision apps. It has $4.3$M parameters and 55 layers.

To illustrate the generality of our method we have tested it on a core set of common benchmark datasets in image classification: CIFAR-10 \cite{cifar}, CIFAR-100 \cite{cifar} and ImageNet \cite{imagenet}. The CIFAR-10 dataset is formed by 60000 $32\times32$ images with 10 classes (6000 images per class, 50000 training images and 10000 test images). CIFAR-100 contains the same number of images as CIFAR-10 but 100 classes (600 images each). On the other hand, Imagenet is a dataset formed by 1431167 annotated images ($224\times224$) and 1000 objects classes. As we have mentioned, OCNNA requires a $D_{var}$ dataset to measure the convolutional filter importance. In the case of CIFAR-10 and CIFAR-100, we have selected $10\%$ of the training images, in other words, 5000 images identically distributed by class. After completing the optimization process, we have evaluated the new model's performance using the test images. For the Imagenet dataset, we have used as $D_{var}$ the "imagenet\_v2/topimages" subset and as test set the "imagenet\_v2/matched-frequency". Both of them have 10000 images sampled after a decade of progress on the original ImageNet dataset, making the new test data independent of existing models and guaranteeing that the accuracy scores are not affected by adaptive overfitting \cite{recht2019imagenet}. These datasets can be found in \cite{imagenetv2}.

\subsection{Metrics}

We measured the prediction performance of the optimized models with accuracy (ACC). In addition, we registered the number of parameters to assess the complexity and the efficiency in terms of memory requirements and runtime (the lower the number of parameters, the higher the efficiency). We also recorded the remaining parameters ratio (RPR) compared with the original model for compression, as other state-of-art approaches carry out. A higher parameter-reduction ration means a smaller model size and, as a result, a less complex model. The definition of RPR is

\begin{equation}
	RPR = 1- \frac{NP_O-NP_S}{NP_O} \label{rpr}
\end{equation}

where $NP_O$ and $NP_S$ represent the number of parameters of the original model and the optimized one, respectively. In any case, we have adapted the metrics used (and the way we show them) to those employed in the state-of-art references to achieve an accurate comparison between OCNNA and the other approaches.

\subsection{Compared state-of-art approaches}

To demonstrate that OCNNA can accurately assess the importance of convolutional filters, we have compared it with other state-of-art approaches for CIFAR-10, CIFAR-100 and ImageNet (Table \ref{tab:state-of-art}). In general terms, all of these methods propose effective and innovative pruning criteria with great results.

\begin{table*}[!t]
	\caption{State-of-art approaches compared to OCNNA divided by dataset \label{tab:state-of-art}}
	\centering
	\begin{tabular}{|c|c|}

		Dataset & Method  \\

		CIFAR-10 & FPEI \cite{WANG202141} \\
		& LRP \cite{https://doi.org/10.48550/arxiv.2002.11018} \\
		& ThiNet \cite{https://doi.org/10.48550/arxiv.1607.03250} \\
		& PScratch \cite{8689357} \\
		& HRank \cite{https://doi.org/10.48550/arxiv.2002.10179} \\
		& Slimming \cite{8237560} \\
		& COP \cite{WANG2021533} \\
		& SOCKS \cite{9755967} \\
		& DeepPruningES \cite{FERNANDESJR202129} \\
		
		\\
		
		CIFAR-100 & FPEI \\
		& SFP \cite{10.5555/3304889.3304970} \\
		& FPGM \cite{10.5555/3304889.3304970} \\
		& HRank \\
		& COP \\
		& Slimming \\
		& DeepPruningES \\
		&  \\
		Imagenet & HRank \cite{https://doi.org/10.48550/arxiv.2002.10179} \\
		& FPEI  \\
		& FPGM  \\
		& SFP  \\
		& PScratch   \\
		& HRank \\
		& $SCWC$ \cite{LI202282} \\
		& RED++ \cite{9786782} \\
		& APoZ \cite{https://doi.org/10.48550/arxiv.1607.03250}  \\
		& CP \cite{https://doi.org/10.48550/arxiv.1707.06168}  \\
		& ThiNet-GAP \cite{https://doi.org/10.48550/arxiv.1607.03250}  \\
		& SSR \cite{8689357}  \\
		& COP \cite{WANG2021533} \\
		& AdaptDCP \cite{9384353} \\
		& ResNet50-pruned-50 and ResNet50-pruned-70 \cite{ALQAHTANI2021103220} \\
		& IoT approach \cite{qi2021} \\
		& LSTM-SEP \cite{9258919} \\
		& AOFP-C2 \cite{9258919} \\	
		& SNACS \cite{9866022}	\\

	\end{tabular}
\end{table*}

\subsection{Training process settings}

For VGG-16, ResNet-50, MobileNet and DenseNet models training on CIFAR-10 or CIFAR-100, we set weight decay as $1e-6$, momentum as $0.9$, learning rate as $1e-3$ and batch size to 64. All images are augmented by horizontal and vertical flip, zoom with range between $0.85$ and $1.5$, rotation range of $180$ and fill mode as reflect, in other words, pixels outside the boundaries of the input image are filled according to the following mode \cite{IDG}:

$$abcddcba|abcd|dcbaabcd$$

We have not trained any model on Imagenet due to the existence  of public available pre-trained models.

\subsection{Results and analysis on CIFAR datasets}

The results on CIFAR are shown in Table \ref{tab:cifar-resnet50}, Table \ref{tab:cifar-vgg16} and Table \ref{tab:cifar-densenet40}. Dataset column shows the learning task; Architecture column shows the neural network used to face the learning task; Base(\%) column refers to the accuracy originally obtained with the before-mentioned architecture for the dataset given; Acc(\%) is the accuracy obtained after applying the pruning algorithm; RPR means the remaining parameters ratio. The lower the better. Acc. Drop is the accuracy loss after pruning. The smaller the better.  As we have said, we used three benchmark architectures, VGG-16, ResNet-50 and DenseNet-40. OCNNA obtains higher accuracy and a smaller RPR than the other state-of-art approaches. 

Given the fact that VGG-16 is a very complex network, it might present a higher redundancy than other architectures. In fact, we are able to reduce the model, pruning the $86.68\%$ parameters for CIFAR-10  and $74.03\%$ for CIFAR-100 (from $138.4$M to $34.6$M parameters) without any accuracy loss for CIFAR-10 (improving in a $0.42\%$) and keeping it nearly unchanged for CIFAR-100 ($-0.44\%$).

In the case of ResNet-50, OCNNA generates a compressed model where just over $45\%$ of the parameters remain ($57.12\%$ for CIFAR-10 and $46.95\%$ for CIFAR-100) while the accuracy loss is very small. As a result, OCNNA is an effective method for compressing CNNs.

For DenseNet-40, OCNNA improves test accuracy for CIFAR-10 ($0.39\%$) and for CIFAR-100 ($0.23\%$) generating a noticeable reduction in the number of parameters, keeping only $52.96\%$ of the units for CIFAR-10 and $34.12\%$ for CIFAR-100.

Finally, for MobileNet, OCNNA improves test accuracy for CIFAR-10 ($0.65\%$) and reduces the error by $0.7\%$ compared with COP v2-0.30, obtaining approximately a $75\%$ reduction in the number of parameters.

\begin{table*}[!t]
	\caption{Results of pruning ResNet-50 on CIFAR datasets. RPR means the remaining parameters ratio. The lower the better. Acc. Drop is the accuracy loss after pruning. The smaller the better. $\dagger$ indicates the best result within the column. \label{tab:cifar-resnet50}}
	\centering
	\resizebox{\textwidth}{!}{
	\begin{tabular}{|c|c|c|c|c|c|c|}

		Dataset & Architecture & Base(\%) & Algorithm & Acc. (\%) &  Acc. Drop (\%) & RPR (\%) \\

		CIFAR-10 & ResNet50 & $93.55$ & FPEI \cite{WANG202141} & $91.85$ & $1.7$ & $45.69$ \\
		&          &         & LRP \cite{https://doi.org/10.48550/arxiv.2002.11018} & $93.37$ & $0.18$ & $75.24$ \\
		&          &         & DeepPruningES (heavy) \cite{FERNANDESJR202129} & $91.89$ & $1.66$ & $78.69$ \\
		&          &         & OCNNA & $93.42$ & $0.13\dagger$ & $42.88\dagger$ \\
		&&&&& \\
		CIFAR-100 & ResNet50 & $73.24$ & FPEI & $69.58$ & $3.66$ & $57.53$ \\
		&          &         & DeepPruningES (heavy) & $57.81$ & $15.43$    & $80.91$ \\
		&          &         & OCNNA & $70.32$ & $2.92 \dagger$    & $53.05 \dagger$ \\

	\end{tabular}}
\end{table*}

\begin{table*}[!t]
	\caption{Results of pruning VGG-16 on CIFAR datasets. RPR means the remaining parameters ratio. The lower the better. Acc. Drop is the accuracy loss after pruning. The smaller the better. $\dagger$ indicates the best result within the column. \label{tab:cifar-vgg16}}
	\centering
	\resizebox{\textwidth}{!}{
	\begin{tabular}{|c|c|c|c|c|c|c|}
		Dataset & Architecture & Base(\%) & Algorithm & Acc. (\%) &  Acc. Drop (\%) & RPR (\%) \\
		CIFAR-10 & VGG-16 & $93.70$ & FPGM \cite{https://doi.org/10.48550/arxiv.1607.03250} & $93.00$ & $0.7$ & $48.97$ \\
		&          &         & DeepPruningES (heavy)  & $91.79$ & $1.91$ & $64.99$ \\
		&          &         & ThiNet \cite{https://doi.org/10.48550/arxiv.1607.03250} & $92.99$ & $0.71$ & $26.92$ \\
		&          &         & PScratch \cite{8689357} & $93.02$ & $0.68$ & $26.96$ \\
		&          &         & HRank \cite{https://doi.org/10.48550/arxiv.2002.10179}  & $93.43$ & $0.27$ & $17.1$ \\
		&          &         & Slimming \cite{8237560} & $93.44$ & $0.26$ & $16.71$ \\
		&          &         & COP v1 \cite{WANG2021533}  & $93.37$ & $0.18$ & $15.15$ \\
		&          &         & COP v2  & $93.86$ & $-0.17$ & $13.56$ \\.
		&          &         & SNACS \cite{9866022} & $91.06$ & $-0.17$ & $3.84 \dagger$ \\
		&          &         & White-Box \cite{9712474}  & $93.47$ & $0.23$ & - \\
		&          &         & SOKS-80\% \cite{9755967}  & $94.01$ & $-0.31$ & - \\
		&          &         & OCNNA  & $94.12 \dagger$ & $-0.42 \dagger$ & $13.32 $ \\
		
		&&& \\
		CIFAR-100 & VGG-16 & $73.51$ & SFP \cite{https://doi.org/10.48550/arxiv.1707.06168} & $71.74$ & $1.77$ & $60.66$ \\
		&          &         & DeepPruningES (heavy)  & $67.06$ & $6.45$ & $80.07$ \\
		&          &         & FPGM   & $72.76$ & $1.25$ & $48.99$ \\
		&          &         & HRank \cite{https://doi.org/10.48550/arxiv.2002.10179}  & $72.43$ & $1.08$ & $44.07$ \\
		&          &         & COP v1  & $72.63$ & $0.88$ & $34.81$ \\
		&          &         & Slimming  & $72.76$ & $0.75$ & $33.4$ \\
		&          &         & COP v2  & $72.99$ & $0.52$ & $26.16$ \\
		&          &         & OCNNA & $73.07 \dagger$ & $0.44 \dagger$    & $25.97 \dagger $\\
		
	\end{tabular}}
\end{table*}

\begin{table*}[!t]
	\caption{Results of pruning DenseNet-40 on CIFAR datasets. RPR means the remaining parameters ratio. The lower the better. Acc. Drop is the accuracy loss after pruning. The smaller the better. $\dagger$ indicates the best result within the column. \label{tab:cifar-densenet40}}
	\centering
	\resizebox{\textwidth}{!}{
	\begin{tabular}{|c|c|c|c|c|c|c|}
		Dataset & Architecture & Base(\%) & Algorithm & Acc. (\%) &  Acc. Drop (\%) & RPR (\%) \\
		CIFAR-10 & DenseNet-40 & $76.52$ & HRank  & $75.94$ & $0.58$ & $58.68$ \\
		&          &         & Slimming  & $75.90$ & $0.62$ & $54.28$ \\
		&          &         & COP v1  & $75.53$ & $0.99$ & $56.1$ \\
		&          &         & COP v2  & $76.03$ & $0.49$ & $54.08$ \\
		&          &         & OCNNA  & $76.91 \dagger$ & $-0.39 \dagger$ & $52.96 \dagger$ \\
		
		&&& \\
		CIFAR-100 & DenseNet-40 & $94.84$ & HRank  & $93.68$ & $0.60$ & $39$ \\
		&          &         & Slimming  & $94.35$ & $0.49$ & $34.8$ \\
		&          &         & COP v1  & $94.19$ & $0.65$ & $37.66$ \\
		&          &         & COP v2  & $94.54$ & $0.30$ & $34.8$ \\
		&          &         & OCNNA  & $95.07 \dagger$ & $-0.23 \dagger$ & $34.12\dagger$ \\

	\end{tabular}}
\end{table*}

\begin{table*}[!t]
	\caption{Results of pruning MobileNet on CIFAR datasets. RPR means the remaining parameters ratio. The lower the better. Acc. Drop is the accuracy loss after pruning. The smaller the better. MobileNet-0.75 means that every layer is $75\%$ of the original one. COP-0.50 implies the $50\%$ of filters are maintained. $\dagger$ indicates the best result within the column. \label{tab:cifar-mobilenet}}
	\centering
	\resizebox{\textwidth}{!}{
	\begin{tabular}{|c|c|c|c|c|c|c|}
		Dataset & Architecture & Base(\%) & Algorithm & Acc. (\%) &  Acc. Drop (\%) & RPR (\%) \\
		CIFAR-10 & MobileNet & $94.07$ & MobileNet-0.75  & $93.36$ & $0.71$ & $53.96$ \\
		&          &         & MobileNet-0.50  & $92.84$ & $1.23$ & $25.28$ \\
		&          &         & COP v1-0.50  & $93.59$ & $0.48$ & $34.06$ \\
		&          &         & COP v1-0.30  & $92.97$ & $1.1$ & $25.94$ \\
		&          &         & COP v2-0.50  & $93.89$ & $0.18$ & $32.72$ \\
		&          &         & COP v2-0.30  & $93.47$ & $0.6$ & $25.38$ \\
		&          &         & Adapt-DCP  & $94.57$ & $-0.6$ & $21.74 \dagger$  \\
		&          &         & OCNNA  & $94.72 \dagger$ & $-0.65 \dagger$ & $24.6 $ \\
		
		&&& \\
		CIFAR-100 & MobileNet & $74.94$ & MobileNet-0.75  & $73.99$ & $0.95$ & $53.96$ \\
		&          &         & MobileNet-0.50  & $73.20$ & $1.74$ & $25.28$ \\
		&          &         & COP v1-0.50  & $73.95$ & $0.99$ & $42.5$ \\
		&          &         & COP v1-0.30  & $73.45$ & $1.49$ & $25.35$ \\
		&          &         & COP v2-0.50  & $74.66$ & $0.28$ & $40.85$ \\
		&          &         & COP v2-0.30  & $74.01$ & $0.93$ & $25.42$ \\
		&          &         & OCNNA  & $74.72 \dagger$ & $0.22 \dagger$ & $23.87 \dagger$ \\

	\end{tabular}}
\end{table*}

\subsection{Results and analysis on Imagenet dataset}

The performance of OCNNA and some state-of-art methods for VGG-16, ResNet-50 and MobileNet on the Imagenet dataset are presented in Table \ref{tab:imagenet-vgg16}, Table \ref{tab:imagenet-resnet50} and Table \ref{tab:imagenet-mobilenet}. For VGG-16 (Table \ref{tab:imagenet-vgg16}) OCNNA, compared to $SCWC (s = 0.2)$, generates a more compressed  ($18.9\%$ of parameters remain vs. $19.7\%$ of SCWC) and more accurate ($2.86\%$ accuracy drop vs. $2.98\%$ for a similar range of compression) model. Noteworthily, our method is designed to reduce the number of parameters as much as possible with the least accuracy loss. Other methods such as APoZ, SSR or SCWC ($s \neq 0.2$) can achieve lower accuracy drops than OCNNA but the RPR for the other state-of-art methods is notably higher.

For ResNet-50 (Table \ref{tab:imagenet-resnet50}), there are much more experimentation in the literature. To the best of our knowledge, SCWC ($s=0.5, s=0.4, s=0.3$) is the only method which improves the accuracy (negative accuracy drop) but with an RPR above $65\%$. FPEI-R7 with DR obtains a notable RPR ($37.88\%$), still smaller than OCNNA ($37.44\%$) but the accuracy drop is quite higher ($1.68\%$ vs $0.57\%$ for OCNNA). In \cite{qi2021} (we call it IoT-Qi) we found the best approach compressing ResNet-50 ($33.7\%$). OCNNA is not as effective as IoT-Qi in terms of RPR but the accuracy drop for OCNNA is more than 4 times smaller compared to IoT-Qi ($0.57\%$ OCNNA vs. $2.38\%$ IoT-Qi). In practical scenarios, it is necessary to balance the performance and compression rate according to different computing requirements, energy consumption restrictions \cite{qi2021} and accuracy requisites. 

For MobileNet (Table \ref{tab:imagenet-mobilenet}), OCNNA outperforms the different approaches of the COP method and the direct simplification of the original model.

Finally, these results support the competitiveness of OCNNA producing simplified CNNs with remarkable complexity reduction while retaining the accuracy.

\begin{table*}[!t]
	\caption{Comparison of OCNNA and other methods on Imagenet (VGG-16). RPR means the remaining parameters ratio. The lower the better. Acc. is the test accuracy after pruning. Acc. Drop is the test accuracy loss after pruning. The smaller the better.  $\dagger$ indicates the best result within the column.\label{tab:imagenet-vgg16}}
	\centering
	\resizebox{\textwidth}{!}{
	\begin{tabular}{|c|c|c|c|c|c|}

		Architecture & Base (\%)& Method & Acc (\%) & Acc. Drop (\%) & RPR (\%) \\

		VGG-16 &$74.4\%$ \cite{https://doi.org/10.48550/arxiv.1409.1556}& $SCWC_{(s=0.6)}$ \cite{LI202282} & $73.8$  & $0.6 \dagger$ & $60.5$ \\
		&& $SCWC_{(s=0.5)}$ & $73.2$ & $1.2$ & $50.3$ \\
		&& $SCWC_{(s=0.4)}$ & $73.17$ & $1.23$ & $40.8$ \\
		&& $SCWC_{(s=0.3)}$ & $72.16$ & $1.64$ & $30.6$ \\
		&& $SCWC_{(s=0.2)}$ & $71.42$ & $2.98$ & $19.7$ \\
		&& APoZ \cite{https://doi.org/10.48550/arxiv.1607.03250} & $73.09$ & $1.31$ & $49$ \\
		&& CP \cite{https://doi.org/10.48550/arxiv.1707.06168} & $70.7$ & $3.7$ & $-$ \\
		&& ThiNet-GAP \cite{https://doi.org/10.48550/arxiv.1607.03250} & $72.64$ & $1.76$ & $-$ \\
		&& SSR \cite{8689357} & $72.75$ & $1.65$ & $-$ \\
		&& OCNNA &  $71.54$ & $2.86$ & $18.9 \dagger$ \\
	\end{tabular}}
\end{table*}

\begin{table*}[!t]
	\caption{Comparison of OCNNA and other methods on Imagenet (ResNet-50). RPR means the remaining parameters ratio. The lower the better. Acc. Drop is the accuracy loss after pruning. The smaller the better.  $\dagger$ indicates the best result within the column.\label{tab:imagenet-resnet50}}
	\centering
	\resizebox{\textwidth}{!}{
	\begin{tabular}{|c|c|c|c|c|c|}
		Architecture & Base(\%) & Method  & Acc. (\%) & Acc. Drop (\%) & RPR (\%) \\
		ResNet-50 & $75.3\%$ \cite{https://doi.org/10.48550/arxiv.1512.03385} & HRank-C1  \cite{https://doi.org/10.48550/arxiv.2002.10179} & $74.13$  & $1.17$ & $62.99$ \\
		&& FPEI-R5 \cite{WANG202141} & $74.36$ & $0.94$ & $61.79$ \\
		&& FPEI-R4 with DR & $74.72$& $0.58$ & $44.31$ \\
		&& HRank-C2 &$71.13$& $4.17$ & $53.71$ \\
		&& FPEI-R6 &$72.23$& $3.07$ & $51.53$ \\
		&& FPEI-R7 with DR  &$73.62$& $1.68$ & $37.88$ \\
		&& SFP \cite{https://doi.org/10.48550/arxiv.1707.06168} &$74.18$& $1.12$ & $61.74$ \\
		&& FPGM \cite{https://doi.org/10.48550/arxiv.1607.03250} &$73.54$& $1.76$ & $61.79$ \\
		&& PScratch \cite{8689357} &$74.75$& $0.55$ & $49.95$ \\
		&& COP v2 \cite{WANG2021533} &$74.97$& $0.33$ & $44.79$ \\
		&& $SCWC_{(s=0.5)}$ \cite{LI202282} &$75.52$& $-0.22 \dagger$ & $77.2$ \\
		&& RED++ \cite{9786782} &$75.2$ & $0.1$ & $55$ \\
		&& $SCWC_{(s=0.4)}$ &$75.45$& $-0.15$ & $72.6$ \\
		&& $SCWC_{(s=0.3)}$ &$75.35$& $-0.05$ & $67.9$ \\
		&& $SCWC_{(s=0.2)}$ &$75.23$& $0.07$ & $63.3$ \\
		&& $SCWC_{(s=0.1)}$ &$75.17$& $0.13$ & $58.6$ \\
		&& Thinet-70 \cite{8416559} \cite{8689357} &$74.03$& $1.27$ & $67.1$ \\
		&& SSR \cite{8689357} &$72.47$& $2.83$ & $47.8$ \\
		&& APoZ \cite{https://doi.org/10.48550/arxiv.1607.03250} &$71.83$& $3.47$ & $47.8$ \\
		&& LSTM-SEP \cite{9258919} &$74.4$& $0.9$ & $57$ \\
		&& AOFP-C2 \cite{9258919} &$75.07$& $0.23$ & $-$ \\
		&& Adapt-DCP \cite{9384353} &$74.44$& $0.86$ & $45.1$ \\
		&& ResNet-50-pruned-70 \cite{ALQAHTANI2021103220} &$75.06$& $0.24$ & $70$ \\
		&	& ResNet-50-pruned-50 \cite{ALQAHTANI2021103220} &$73.99$& $1.31$ & $50$ \\
		&& IoT-Qi \cite{qi2021} &$72.92$& $2.38$ & $33.7 \dagger$ \\
		&& SNACS \cite{9866022} & $74.65$ & $0.65$ & $44.9$ \\
		&& White-Box \cite{9712474} & $74.21$ & $1.09$ & - \\
		&& OCNNA  &$74.73$& $0.57$ & $37.44$ \\

	\end{tabular}}
\end{table*}

\begin{table*}[!t]
	\caption{Results of pruning MobileNet on the Imagenet dataset. RPR means the remaining parameters ratio. The lower the better. Acc. Drop is the accuracy loss after pruning. The smaller the better. MobileNet-0.75 means that every layer is $75\%$ of the original one. COP-0.50 implies the $50\%$ of filters are maintained. $\dagger$ indicates the best result within the column. \label{tab:imagenet-mobilenet}}
	\centering
	\resizebox{\textwidth}{!}{
	\begin{tabular}{|c|c|c|c|c|c|c|}
		Dataset & Architecture & Base(\%) & Algorithm & Acc. (\%) &  Acc. Drop (\%) & RPR (\%) \\

		Imagenet & MobileNet & $69.96$ & MobileNet-0.75  & $68.01$ & $1.95$ & $60.94$ \\
		&          &         & MobileNet-0.50  & $63.29$ & $6.67$ & $36.29$ \\
		&          &         & COP v1-0.70  & $68.52$ & $1.44$ & $59.31$ \\
		&          &         & COP v1-0.40  & $64.38$ & $5.58$ & $28.99$ \\
		&          &         & COP v2-0.70  & $69.02$ & $0.94$ & $57.09$ \\
		&          &         & COP v2-0.40  & $65.33$ & $4.63$ & $29.81$ \\
		&          &         & Adapt-DCP  & $69.58$ & $0.38$ & $66.73$ \\
		&          &         & OCNNA  & $69.75 \dagger$ & $0.21 \dagger$ & $27.22 \dagger$ \\
		
	\end{tabular}}
\end{table*}

\section{Ablation study}

\begin{figure*}[!t]
	\centering
	\includegraphics[width=\linewidth]{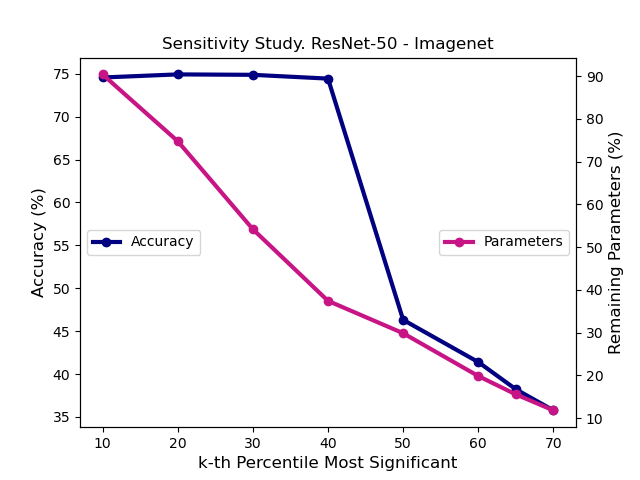}
	\caption{Sensitivity study of $k$ percentile of significance value for ResNet-50 and Imagenet dataset. Left Y-axis shows the Test Accuracy and Right Y-axis shows the remaining parameters ratio. The base accuracy is $75.4\%$. As we can see, when $k=40$ (40-th percentile), we obtain a significant reduction of parameters (remaining $37.44\%$) with an accuracy drop of $0.57\%$.}
	\label{study-k}
\end{figure*}

As we have seen, OCNNA is a parametric algorithm designed to simplify CNN models. It can be applied to any convolutional model, as ResNet or VGG networks, without any adjustment, in contrast with other state-of-art approaches as FPEI \cite{WANG202141} which presents different versions (FPEI, FPEI-R4 with DDR, FPEI-R5, FPEI-R6, FPEI-R7 with DR) in order to ensure the quality in prediction in different situations. OCNNA counts only with one parameter, $k$, which represents the $k$-th percentile of filters with higher importance, after applying transformations such as PCA, Frobenius Norm and CV. How does changing the value of $k$ affect in terms of accuracy and network simplification? It is illustrated through the experiment depicted in Figure \ref{study-k}. In this experiment, different values of $k$, ranging between 10 and 75 have been used and the resulting accuracy and final number of parameters evaluated. As $k$ value increases, OCNNA boosts the reduction of the number of filters which will formed part of the new model. In consequence, the number of parameters substantially drops. Nonetheless, the accuracy also suffers a dramatically drop as $k$ increases. Notice the remarkable drop between $k=40$ ($74.43\%$ in accuracy) and $k=50$ ($46.31\%$). In fact, $40$-th percentile is the best value of $k$, obtaining the highest possible accuracy bearing in mind the significant reduction of parameters.

\section{Conclusions}
Deep Neural Networks have become the state-of-art technique for several AI challenging tasks. In particular, CNNs have achieved and extraordinary success in a wide range of computer vision problems. However, these neural networks entail significant energetic costs and are hard to design efficiently. 

In this paper, we propose OCNNA, a novel CNN optimization and construction method based on pruning and knowledge distillation designed to decide the importance of convolutional layers, ordering the filters (features) by importance. Our proposed strategy can carry out effective end-to-end training and compression of CNNs. It is easy to apply and depends on a single parameter $k$, called percentile of significance, which represents the proportion of filters which will be transferred to the new model based on their importance. Only the $k$-th percentile of filters with higher values after applying the OCNNA process (PCA for feature selection, Frobenius Norm for summary and CV for measuring variability) will form part of the new optimized model. 

The proposal has been evaluated through a thorough empirical study including the best known datasets (CIFAR-10, CIFAR-100 and Imagenet) and CNN architectures (VGG-16, ResNet-50, DenseNet-40 and MobileNet). The experimental results, comparing with 20 state-of-art CNN simplification techniques and obtaining successful results, confirm that simpler CNN models can be obtained with small accuracy losses by distilling knowledge from the original models to the new ones. As a result,  OCNNA is a competitive CNN constructing method based on pruning and knowledge distillation which could ease the deployment of AI models into IoT or resource-limited devices.

\section*{Acknowledgment}
This research has been partially supported by the projects with references TIN2016-81113-R, PID2020-118224RB-100 granted by the Spain's Ministry of Science and Innovation, and the project with reference P18-TP-5168 granted by Industry Andalusian Council (Consejería de Transformación Económica, Industria, Conocimiento y Universidades de la Junta de Andalucía), with the cofinance of the European Regional Development Fund (ERDF).

\end{document}